\documentclass{article}
\usepackage[preprint]{neurips_2026}
\usepackage[utf8]{inputenc} 
\usepackage[T1]{fontenc}    
\usepackage{hyperref}       
\usepackage{url}            
\usepackage{booktabs}       
\usepackage{amsfonts}       
\usepackage{nicefrac}       
\usepackage{microtype}      
\usepackage{xcolor}         
\usepackage{amsmath}
\usepackage{multirow}
\usepackage{graphicx}
\usepackage{amsthm}
\newtheorem{proposition}{Proposition}

\title{Local Truncation Error-Guided Neural ODEs for Large Scale Traffic Forecasting}
\author{
Xiao Zhang \\
Zhengzhou University \\
\texttt{xiaozhang@zzu.edu.cn}\\
\And
Yafei Li \\
Zhengzhou University \\
\texttt{ieyfli@zzu.edu.cn}\\
\And
Ruixiang Wang \\
Zhengzhou University \\
\texttt{wang\_rui\_xiang@gs.zzu.edu.cn}\\
\AND
Wei Wei \\
Northwestern Polytechnical University \\
\texttt{weiweinwpu@nwpu.edu.cn}\\
\And
Shuo He \\
Zhengzhou University \\
\texttt{heshuo@zzu.edu.cn}\\
\And
Mingliang Xu \\
Zhengzhou University \\
\texttt{iexumingliang@zzu.edu.cn}\\
}

\begin{document}
	
	\maketitle
	
	\begin{abstract}
		Spatiotemporal forecasting in physical systems, such as large-scale traffic networks, requires modeling a dual dynamic: continuous macroscopic rhythms and discrete, unpredictable microscopic shocks. While Neural Ordinary Differential Equations (ODEs) excel at capturing smooth evolution, their inherent Lipschitz continuity constraints inevitably cause severe over-smoothing when confronting abrupt anomalies. Recent physics-informed methods attempt to bypass this by penalizing numerical integration errors to enforce manifold smoothness. However, we mathematically reveal that such rigid regularization inherently triggers gradient conflicts and ``attention collapse,'' stripping the model of its sensitivity to anomalies. To resolve this continuity-shock dilemma, we propose Local Truncation Error-Guided Neural ODEs (LTE-ODE). Rather than treating numerical error as a nuisance to be eliminated, we innovatively repurpose the Local Truncation Error (LTE) as an unsupervised forward inductive bias. By mapping the LTE into a dynamic spatial attention mask, our architecture gracefully preserves high-precision continuous ODE evolution in stable regions, while adaptively triggering a discrete compensation branch exclusively at shock points. Trained purely end-to-end without manifold penalties, LTE-ODE achieves state-of-the-art performance on multiple large-scale benchmarks, exhibiting exceptional robustness against highly non-linear fluctuations. Furthermore, our ablation on integration steps demonstrates high deployment flexibility, allowing the model to seamlessly adapt to varying hardware memory constraints in real-world applications.
	\end{abstract}
	
	\section{Introduction}
	\label{sec:intro}
	Spatiotemporal forecasting is a core challenge across numerous real-world applications, particularly in large-scale traffic flow management. In recent years, deep learning methods have made significant strides in this domain. Early dominant paradigms primarily relied on discrete network architectures, utilizing Graph Neural Networks (GNNs) \citep{huang2020lsgcn, zheng2023spatio,li2021spatial,duan2023localised} to capture spatial dependencies, coupled with Recurrent Neural Networks (RNNs)	\citep{yao2018modeling} or Transformers \citep{fang2025efficient, PDFormer, chen2022bidirectional} to process temporal dynamics. Despite their immense success, these discrete models are typically constrained by fixed sampling frequencies, inherently limiting their ability to model the underlying continuous evolutionary processes of physical systems.
	
	To overcome the architectural bottlenecks of discrete models, continuous-time models-most notably Neural Ordinary Differential Equations (Neural ODEs)\citep{chen2018neural,chen2023contiformer,zhang2023dual,liu2025graph,fang2021spatial,zhang2026yourself}-have emerged as a foundational framework. By parameterizing the derivative of hidden states with continuous vector fields, Neural ODEs can elegantly handle irregularly sampled observations and effectively capture the smooth, macroscopic temporal rhythms governing traffic flows. 
	
	However, real-world physical systems exhibit a fundamental ``dual nature'': they follow continuous, predictable periodic rhythms, yet are replete with discrete, highly unpredictable abrupt anomalies (e.g., sudden congestion, or extreme weather). This reality exposes a fatal vulnerability in standard Neural ODEs. Constrained by the Picard-Lindel\"{o}f theorem, standard ODEs require their vector fields to satisfy Lipschitz continuity, meaning that the system's evolutionary trajectories form a homeomorphism and can never intersect. Consequently, when confronted with sudden high-frequency anomalies, these models inevitably suffer from severe ``over-smoothing,'' failing to capture divergent future trajectories that suddenly branch out from otherwise identical historical states.
	
	Recognizing this critical limitation, recent physics-informed deep learning models have attempted to constrain continuous evolution by penalizing numerical integration errors. A common practice is to minimize the numerical discrepancy between solvers of different orders to enforce manifold smoothness. In this paper, we argue that this approach harbors a fundamental mathematical flaw. We reveal-both theoretically and empirically-that such rigid manifold regularization intrinsically triggers severe gradient conflicts. When the model requires a large numerical deviation to escape the smooth continuous manifold at a point of sudden shock, the loss function's penalty on the error arbitrarily suppresses the model's sensitivity to anomalies. We rigorously formalize this catastrophic phenomenon as ``attention collapse.''
	
	To break the modeling dilemma between continuous rhythms and discrete mutations, we propose a novel paradigm: Local Truncation Error-Guided Neural ODEs (LTE-ODE). Rather than treating numerical error as a nuisance to be eliminated, we innovatively repurpose the Local Truncation Error (LTE)-the theoretical discrepancy between a first-order Euler trajectory and a second-order Runge-Kutta trajectory-into an unsupervised forward inductive bias. By dynamically mapping the LTE into a spatial attention mask, LTE-ODE seamlessly preserves high-precision continuous evolution in smooth regions while adaptively triggering a discrete deep feature compensation branch exclusively at localized shock points. Optimized purely in a data-driven, end-to-end manner, the network learns to ``exploit'' numerical discrepancies rather than ``penalize'' them.
	
	The main contributions of this paper are summarized as follows:
	\begin{itemize}
		\item \textbf{Architectural Innovation and Expressivity:} We propose LTE-ODE, a novel continuous-discrete hybrid dynamical system. We theoretically prove that the LTE-guided discrete compensator breaks the homeomorphic constraints of standard ODEs, fundamentally expanding the model's topological representational capacity to capture divergent, highly non-linear trajectories.
		
		\item \textbf{Theoretical Formalization of Attention Collapse:} Based on gradient flow analysis, we provide a rigorous mathematical proof demonstrating that penalizing physical numerical errors inevitably leads to ``attention collapse,'' compromising generalization ability. This provides a solid theoretical foundation for our purely data-driven optimization strategy.
		
		\item \textbf{High-Efficiency Computation via Embedded Runge-Kutta:} Our embedded integration scheme requires strictly zero additional Neural Function Evaluations (NFEs) for error estimation. Exploiting spatial sparsity further restricts discrete computations to anomalous nodes, making LTE-ODE highly scalable to massive graphs.
		
		\item \textbf{State-of-the-Art Performance and Deployment Flexibility:} Extensive experiments on large-scale benchmarks indicate that LTE-ODE outperforms existing baselines. Furthermore, our ablation studies elevate the ODE integration step from a simple hyperparameter to a strategic tuning knob, demonstrating the model's flexible deployment capabilities across varying hardware constraints (e.g., edge devices vs. cloud computing).
	\end{itemize}
	
	\section{The LTE-ODE Framework}
	\label{method}
	
	To successfully resolve the fundamental modeling conflict between ``smooth evolution'' and ``discrete abrupt shocks'' in spatiotemporal dynamics, we propose the Local Truncation Error-Guided Neural Ordinary Differential Equation (LTE-ODE). 
	
\begin{figure}[htbp]
		\centering
		\includegraphics[width=0.95\linewidth]{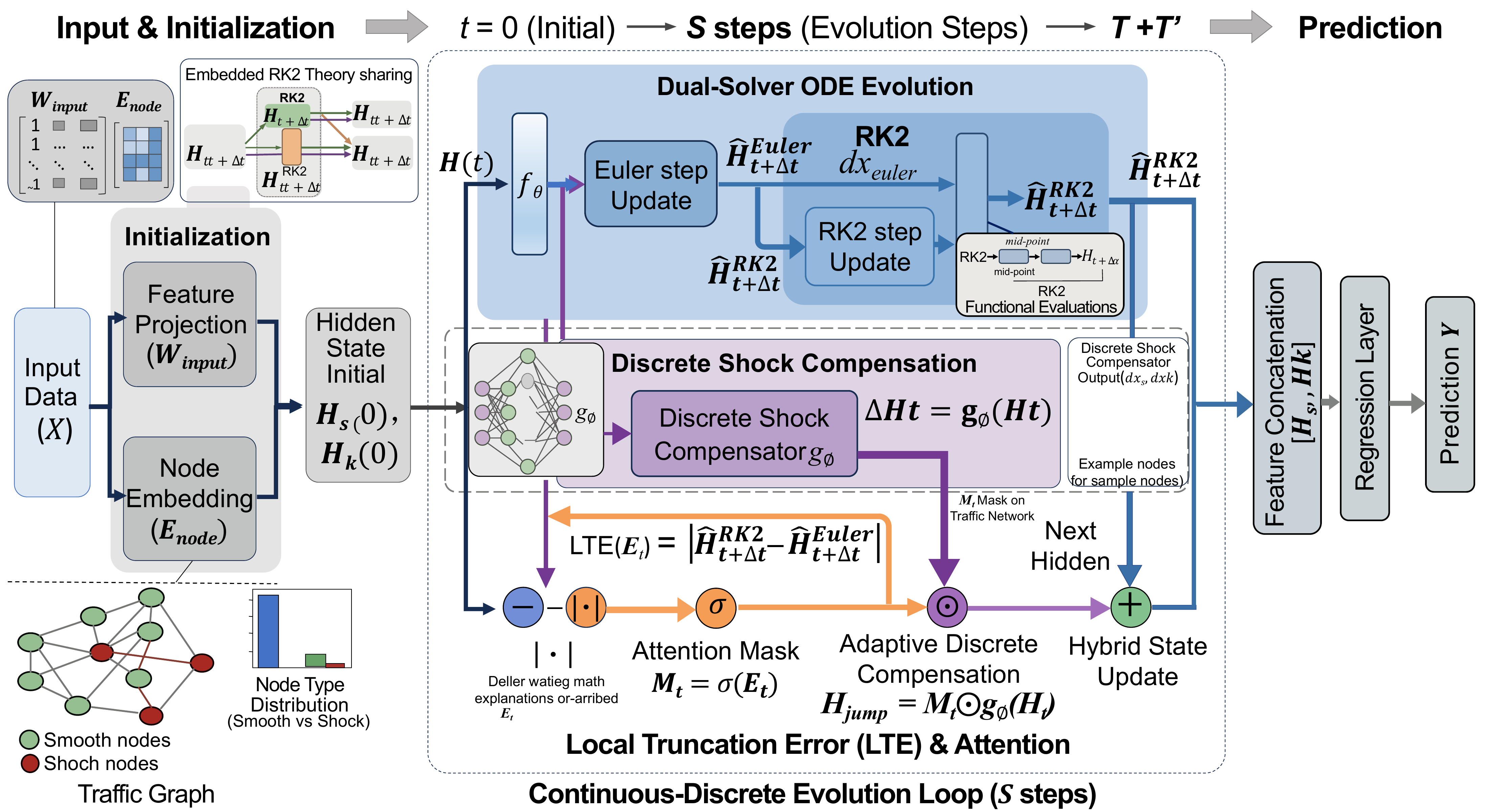}
		\caption{The overall framework of the proposed LTE-ODE. \textbf{(Left) Input \& Initialization:} The input data $\mathbf{X}$ is projected via $\mathbf{W}_{input}$ and concatenated with node embeddings $\mathbf{E}_{node}$ to form initial dual-stream states. \textbf{(Middle) Continuous-Discrete Evolution Loop:} Over $S$ steps, a dual-solver (Euler and RK2) evaluates the continuous vector field $f_\theta$. The numerical discrepancy between them yields the Local Truncation Error (LTE) $\mathbf{E}_t$, which generates a spatial attention mask $\mathbf{M}_t$. This mask dynamically activates the discrete shock compensator $g_\phi$ to inject high-frequency momentum into the continuous trajectory. \textbf{(Right) Prediction:} Evolved features are concatenated and passed through a regression layer. \textbf{Key Take-away:} The LTE serves as an unsupervised, physics-informed bridge—it gracefully maintains pure continuous ODE evolution in smooth traffic regions while forcefully activating discrete compensations specifically at nodes experiencing abrupt shocks.}
		\label{fig:framework}
	\end{figure}

	The overall architecture of our proposed framework is illustrated in Figure \ref{fig:framework}. Rather than relying on a static architecture, LTE-ODE operates as a dynamically coupled system consisting of three phases: 1) \textit{State Initialization}, which embeds raw spatiotemporal signals into a robust dual-stream latent space; 2) \textit{Continuous-Discrete Evolution}, our core contribution, which intertwines standard ODE integration with adaptive discrete jumps; and 3) \textit{Horizon Forecasting}, which decodes the evolved hybrid states into future predictions. In the following subsections, we explicitly detail how this architecture naturally absorbs macroscopic rhythms.
	
	\subsection{Dual-Stream Continuous ODE Backbone}
	\label{subsec:problem_formulation}
	
	Spatiotemporal sequence forecasting aims to deduce future system states based on historical signals. Given a traffic graph $\mathcal{G}$ comprising $N$ nodes and observed signals $\mathbf{X} \in \mathbb{R}^{B \times N \times T \times D}$ over the past $T$ time steps (where $B$ is the batch size and $D$ is the feature dimension), the objective is to learn a mapping to predict future signals $\mathbf{Y} \in \mathbb{R}^{B \times N \times T' \times 1}$ over the subsequent $T'$ steps.
	
	To comprehensively capture node heterogeneity and complex spatiotemporal interactions, our model initially projects the input signal $\mathbf{X}$ into a high-dimensional latent space via a linear projection weight matrix $\mathbf{W}_{input}$. This projected feature is then concatenated with learnable node embeddings $\mathbf{E}_{node}$ along the channel dimension. This initialization constructs a dual-stream latent state: $\mathbf{H}_s(0)$ representing local spatial dynamics, and $\mathbf{H}_k(0)$ representing global topological dynamics. In the subsequent continuous evolution, we unify the notation as $\mathbf{H}(t)$ to denote the latent state of either stream at time $t$.
	
	The continuous-time evolution of the hidden state $\mathbf{H}(t)$ is governed by a neural network-parameterized ordinary differential equation:
	\begin{equation}
		\frac{d\mathbf{H}(t)}{dt} = f_\theta(\mathbf{H}(t), \mathcal{G})
	\end{equation}
	where $\theta$ denotes the learnable parameters of the continuous vector field $f_\theta$, implemented as a weight-sharing linearized convolution. During the actual forward propagation, to actively perceive the curvature changes of the dynamical manifold within a given integration step $\Delta t$, we compute a first-order coarse estimation (Euler method) and a second-order refined estimation (Runge-Kutta 2 method) in parallel:
	\begin{equation}
		\mathbf{H}_{t+\Delta t}^{\text{Euler}} = \mathbf{H}_t + \Delta t \cdot f_\theta(\mathbf{H}_t, \mathcal{G})
	\end{equation}
	\begin{equation}
		\mathbf{H}_{t+\Delta t}^{\text{RK2}} = \mathbf{H}_t + \Delta t \cdot f_\theta \left( \mathbf{H}_t + \frac{\Delta t}{2} f_\theta(\mathbf{H}_t, \mathcal{G}), \mathcal{G} \right)
	\end{equation}
	Because both methods inherently share the internal intermediate evaluation states (i.e., the embedded property of Runge-Kutta solvers), this dual-solver mechanism efficiently generates two continuous evolutionary trajectories with different orders of truncation error while incurring almost zero additional vector field evaluation overhead.
	
	\subsection{LTE-Guided Adaptive Discrete Shock Compensation}
	\label{subsec:lte_compensation}
	
	Standard ODEs, strictly constrained by Lipschitz continuity, inherently fail to fit discontinuous, abrupt shocks in traffic flows. When the underlying system encounters drastic state jumps, the predictive trajectories of the low-order (Euler) and high-order (RK2) solvers diverge significantly due to their differing truncation error orders. We define this physical discrepancy as the Local Truncation Error (LTE), denoted as $\mathbf{E}_t$:
	\begin{equation}
		\mathbf{E}_t = \left| \mathbf{H}_{t+\Delta t}^{\text{RK2}} - \mathbf{H}_{t+\Delta t}^{\text{Euler}} \right|
	\end{equation}
	Unlike prior works that treat $\mathbf{E}_t$ merely as numerical noise to be penalized, we innovatively repurpose it as an unsupervised forward inductive bias. From a dynamical systems perspective, a larger absolute value of $\mathbf{E}_t$ at a specific node indicates that its current behavior severely deviates from the smooth manifold. Therefore, we generate a spatially adaptive attention mask $\mathbf{M}_t \in (0, 1)$ via a non-linear activation function:
	\begin{equation}
		\mathbf{M}_t = \sigma(\mathbf{E}_t)
	\end{equation}
	where $\sigma(\cdot)$ represents the Sigmoid activation function, bounding the discrepancy signals into a normalized gating mechanism.
	
	To explicitly break the topological constraints of continuous evolution, we introduce a discrete feature compensation operator $g_\phi(\cdot)$, parameterized by $\phi$, that is strictly free from time-shared weight constraints. This operator is specifically designed to capture the high-frequency discrete jumps that the continuous vector field structurally cannot fit. Utilizing the mask $\mathbf{M}_t$ as a localized spatial gating signal, the final hybrid hidden state update equation at time $t+\Delta t$ is formulated as:
	\begin{equation}
		\mathbf{H}_{t+\Delta t} = \mathbf{H}_{t+\Delta t}^{\text{RK2}} + \mathbf{M}_t \odot g_\phi(\mathbf{H}_t)
	\end{equation}
	where $\odot$ denotes the element-wise Hadamard product. 	
	This hybrid evolutionary mechanism ensures maximum adaptability and spatial sparsity. In stable traffic regions, the numerical error $\mathbf{E}_t \to 0$, thereby shutting off the mask ($\mathbf{M}_t \to 0$); the model gracefully degenerates into a pure, high-precision continuous ODE, ensuring the smooth prediction of macroscopic rhythms. Conversely, at nodes experiencing accidents or sudden congestion, the solver discrepancy spikes ($\mathbf{E}_t \gg 0$), highly activating the mask ($\mathbf{M}_t \to 1$). The discrete compensator $g_\phi(\cdot)$ then forcefully injects spatial jump momentum into the system, allowing the state to instantly traverse the manifold and achieve precise capture of extreme fluctuations.
	
	\subsection{End-to-End Optimization and the ``No-Regularization'' Paradigm}
	\label{subsec:optimization}
	
	After continuous propagation over a predefined number of integration steps $S$, the model extracts and concatenates the dual-stream features, mapping them through a Regression Layer to yield the final prediction tensor $\mathbf{\hat{Y}}$. We employ the Mean Absolute Error (MAE) as the loss function:
	\begin{equation}
		\mathcal{L}_{\text{task}} = \frac{1}{B \times N \times T'} \sum \left| \mathbf{\hat{Y}} - \mathbf{Y} \right|
	\end{equation}	
	Crucially, LTE-ODE adopts an optimization philosophy that fundamentally contrasts with existing physics-informed models. State-of-the-art methods (e.g., ODE variants incorporating physical constraints) typically assume that a large $\mathbf{E}_t$ disrupts system stability, and thus forcibly introduce manifold regularization terms (e.g., $\mathcal{L} = \mathcal{L}_{\text{task}} + \lambda \|\mathbf{E}_t\|$) into the loss function to smooth the trajectories.
	
	In stark contrast, LTE-ODE operates under a strict ``no-regularization'' paradigm, driven solely by the end-to-end task loss ($\mathcal{L}_{\text{task}}$). To accurately capture abrupt shocks, the optimizer naturally amplifies the localized numerical error $\mathbf{E}_t$ to fully activate the discrete compensator. Imposing a manifold penalty on $\mathbf{E}_t$ creates an irreconcilable gradient conflict. As proven in Section \ref{theory}, this forced smoothing inevitably induces ``Attention Collapse,'' stripping the model of its ability to resolve highly non-linear anomalies.
			
	\section{Mechanistic Analysis}
	\label{theory}
	
	In the previous section, we introduced the hybrid continuous-discrete architecture of LTE-ODE. Here, rather than presenting isolated mathematical theories, we delve into the underlying mechanics of our design choices. We analyze how LTE-ODE structurally overcomes topological bottlenecks and why our strict ``no-regularization'' paradigm is crucial for preserving spatial selectivity. 
	
	\subsection{Expressivity and Topology: Breaking the Picard-Lindel\"{o}f Constraints}
	\label{sec:topology}
	
	A fundamental bottleneck of standard Neural ODEs in spatiotemporal forecasting is their theoretical upper bound in modeling divergent anomalies. 
	
	\begin{proposition}[Breaking Topological Constraints]
		\label{prop:topology}
		Standard Neural ODEs preserve the topology of the input space. LTE-ODE transforms the system into a piecewise-continuous dynamical system via the discrete compensation term $\mathbf{M}_t \odot g_\phi(\mathbf{H}_t)$, permitting trajectory intersections and fundamentally expanding topological expressivity. (Formal proof is provided in Appendix \ref{app:proof_topology}).
	\end{proposition}
	
	\textbf{Mechanistic Insight:} In physical terms, if two traffic nodes share identical historical states, standard ODEs (bound by Lipschitz continuity) are mathematically forced to predict identical smooth futures for both. By using the numerical solver discrepancy ($\mathbf{E}_t$) to trigger a discrete jump, LTE-ODE instantly breaks this homeomorphic assumption. This allows the model to map one node to a smooth trajectory while forcefully routing the other to a divergent, ``abrupt shock'' phase space, aligning perfectly with the dual nature of real-world traffic.
	
	\subsection{Optimization Dynamics: Mitigating ``Attention Collapse''}
	\label{sec:gradient_flow}
	
	Previous physics-informed spatiotemporal models intuitively assume that numerical integration errors undermine system stability, prompting them to introduce manifold regularization penalties (e.g., $\mathcal{L}_{\text{total}} = \mathcal{L}_{\text{task}} + \lambda \|\mathbf{E}_t\|$).
	
	\begin{proposition}[Attention Collapse Theorem]
		\label{prop:collapse}
		Enforcing a regularization constraint that penalizes $\mathbf{E}_t \to 0$ induces severe gradient conflicts, degrading the attention mask $\mathbf{M}_t$ into a non-informative constant and triggering attention collapse. (Formal proof is provided in Appendix \ref{app:proof_collapse}).
	\end{proposition}
	
	\textbf{Mechanistic Insight:} This theorem explains the pathology of traditional regularization. When the primary forecasting task demands a discrete jump to fit an anomaly, it actively tries to maximize the mask activation. If we simultaneously apply a manifold penalty forcing the error to zero, we create an orthogonal gradient conflict. The optimizer compromises by collapsing the mask to a uniform $0.5$ across all nodes. By strictly adopting a pure, end-to-end task-driven loss, LTE-ODE preserves the sharp, heavy-tail distribution of the attention mask, ensuring the discrete compensator acts as a highly selective anomaly detector rather than uninformative spatial noise.
	
	\subsection{Computational Efficiency: Embedded Solvers and Spatial Sparsity}
	\label{sec:complexity}
	
	A common critique of hybrid or multi-order ODE solvers is the perceived inflation of computational operations and memory footprint. However, LTE-ODE bypasses this bottleneck through algorithmic embedding and conditional computation.
	
	\begin{proposition}[Computational Complexity Bounds]
		\label{prop:complexity}
		LTE-ODE incurs strictly zero additional Neural Function Evaluations (NFEs) for truncation error estimation, maintaining a time complexity of $\mathcal{O}(2C)$ (where $C$ is the cost of a single vector field evaluation), equivalent to a standalone RK2 solver. Furthermore, its spatial sparsity strictly bounds the dense matrix operations of the discrete compensator. (Detailed complexity analysis is provided in Appendix \ref{app:proof_complexity}).
	\end{proposition}
	
	\textbf{Mechanistic Insight:} Instead of running the neural network multiple independent times to estimate numerical discrepancies, LTE-ODE is designed akin to an Embedded Runge-Kutta method. The first-order Euler step is naturally generated as a mathematically ``free'' intermediate byproduct while computing the second-order RK2 step. Moreover, because the truncation error $\mathbf{E}_t \to 0$ for the vast majority of stable traffic nodes, the mask acts as a conditional computation gate. The heavy discrete compensator $g_\phi(\cdot)$ is effectively executed only on a highly sparse subset of anomalous nodes, allowing the architecture to scale effortlessly to massive spatial graphs.
	
	\section{Experiments}
	\label{sec:experiments}
	
	\subsection{Experimental Settings}
	\label{subsec:details}
		To evaluate the effectiveness of the proposed LTE-ODE, we conducted extensive experiments on four real-world traffic forecasting datasets. Following the rigorous experimental protocol established by GSNet~\citep{GSNet}, we benchmark our model against 13 state-of-the-art baselines to demonstrate its superior predictive performance and adaptability. Due to space constraints, detailed descriptions of experimental details are provided in Appendix \ref{app:exp_details}.
	
	\textbf{Datasets and Metrics.} We evaluate LTE-ODE on four widely-used public spatiotemporal benchmarks\citep{BigST,GSNet,PDFormer}: PEMS07, PEMS08, CA, and England. Forecasting performance is measured using Mean Absolute Error (MAE), Mean Absolute Percentage Error (MAPE), and Root Mean Squared Error (RMSE). 
	
	\textbf{Baselines.} We compare our method against a comprehensive suite of 13 baselines, ranging from traditional statistical models (ARIMA\citep{williams2003modeling}) to recent deep learning architectures, including spatial-temporal GNNs (e.g., GWNet~\citep{wu2019graph}, AGCRN~\citep{bai2020adaptive}) and advanced Transformer-based models (e.g., PDFormer~\citep{PDFormer}, UniST~\citep{UniST}, GSNet~\citep{GSNet}).
	
	\subsection{Main Results}
	\label{subsec:results}
	
	Table~\ref{tab:mainresults} summarizes the performance of all models across the four datasets, averaged over the 12 forecasting horizons.
	
	\vspace{-0.4cm}
	\begin{table}[!htbp]
		\caption{Performance comparison on four datasets.}
		\label{tab:mainresults}
		\vspace{-0.2cm}
		\centering
		\resizebox{\linewidth}{!}{
			\begin{tabular}{l|ccc|ccc|ccc|ccc}
				\toprule
				\multirow{2}{*}{Dataset} 
				& \multicolumn{3}{c|}{PEMS08} 
				& \multicolumn{3}{c|}{England} 
				& \multicolumn{3}{c|}{PEMS07} 
				& \multicolumn{3}{c}{CA} \\
				\cmidrule(lr){2-4} \cmidrule(lr){5-7} \cmidrule(lr){8-10} \cmidrule(lr){11-13}
				Metric 
				& MAE & MAPE & RMSE 
				& MAE & MAPE & RMSE 
				& MAE & MAPE & RMSE 
				& MAE & MAPE & RMSE \\
				\midrule
				ARIMA   & 31.23 & 19.25 & 33.47 & 4.23 & 5.72 & 7.68 & 33.89 & 17.60 & 46.38 & 34.26 & 21.35 & 44.68 \\
				STResNet & 23.25 & 15.58 & 32.25 & 4.03 & 5.68 & 7.55 & 29.36 & 15.24 & 42.46 & 29.25 & 20.33 & 40.63 \\
				ACFM    & 15.86 & 10.13 & 25.34 & 3.52 & 5.28 & 7.86 & 25.86 & 11.83 & 39.03 & 26.38 & 19.24 & 37.86 \\
				STGCN   & 17.50 & 11.29 & 27.03 & 3.55 & 5.30 & 7.38 & 25.32 & 11.16 & 39.27 & 25.68 & 18.43 & 36.44 \\
				DCRNN   & 17.86 & 11.45 & 27.83 & 3.59 & 4.90 & 7.42 & 25.30 & 11.66 & 38.58 & 25.75 & 18.62 & 36.91 \\
				GWNet   & 19.13 & 12.68 & 31.05 & 3.53 & 4.93 & 7.57 & 26.85 & 12.12 & 42.78 & 24.73 & 17.46 & 35.86 \\
				AGCRN   & 15.95 & 10.09 & 25.22 & 3.32 & 5.01 & 7.33 & 22.37 & 9.12 & 36.55 & 25.03 & 17.93 & 36.23 \\
				PDFormer& \textbf{13.58} & \textbf{9.05} & 23.51 & 3.46 & 4.97 & 7.46 & 20.42 & 8.86 & 32.87 & 25.22 & 18.96 & 35.43 \\
				Bi-STAT & 13.62 & 9.43 & 23.17 & 3.42 & 4.89 & 7.54 & 21.13 & 8.98 & 33.86 & 25.36 & 19.03 & 35.27 \\
				GPT-ST  & 14.85 & 9.63 & 24.32 & 3.43 & 4.97 & 7.42 & 21.43 & 9.32 & 34.76 & 24.93 & 17.52 & 35.86 \\
				UniST   & 16.01 & 10.23 & 25.37 & 3.49 & 4.88 & 7.51 & 24.62 & 11.56 & 38.21 & 25.53 & 18.87 & 35.46 \\
				AGS     & 15.50 & 9.71 & 25.01 & 3.15 & 4.56 & 7.52 & 21.56 & 9.03 & 34.90 & 24.96 & 17.88 & 36.02 \\
				BigST   & 16.35 & 10.05 & 24.37 & 3.63 & 5.13 & 7.76 & 22.86 & 10.02 & 34.03 & 23.36 & 16.81 & 34.98 \\
				GSNet   & 14.76 & 9.48 & 22.52 & 3.20 & 4.82 & 6.60 & 20.70 & 8.77 & 32.66 & 19.76 & 14.39& 30.99\\
				\midrule
				LTE-ODE & 14.56 & 9.41 & \textbf{22.15} & \textbf{2.95} & \textbf{4.27} & \textbf{6.25} & \textbf{20.20} & \textbf{8.61} & \textbf{31.98} & \textbf{18.73} & \textbf{13.53} & \textbf{28.89} \\
				\bottomrule
		\end{tabular}}
	\end{table}
	\vspace{-0.3cm}
	
	As shown in Table~\ref{tab:mainresults}, LTE-ODE achieves state-of-the-art performance across nearly all metrics on the evaluated datasets. Specifically, LTE-ODE demonstrates significant improvements in RMSE (e.g., $28.89$ on CA, a substantial reduction compared to the best baselines). Because RMSE heavily penalizes large variance errors, this robust improvement empirically validates our theoretical claim: LTE-ODE is uniquely capable of accurately capturing and modeling sudden, high-amplitude traffic shocks that standard continuous models structurally fail to fit. Even when compared against recent computationally heavy Transformer-based models like PDFormer and UniST, LTE-ODE exhibits superior generalization and predictive stability.
	
	\subsection{Computational Efficiency and Scalability}
	\label{subsec:efficiency}
	
\begin{figure}[!htbp]
		\centering
		\includegraphics[width=0.95\linewidth]{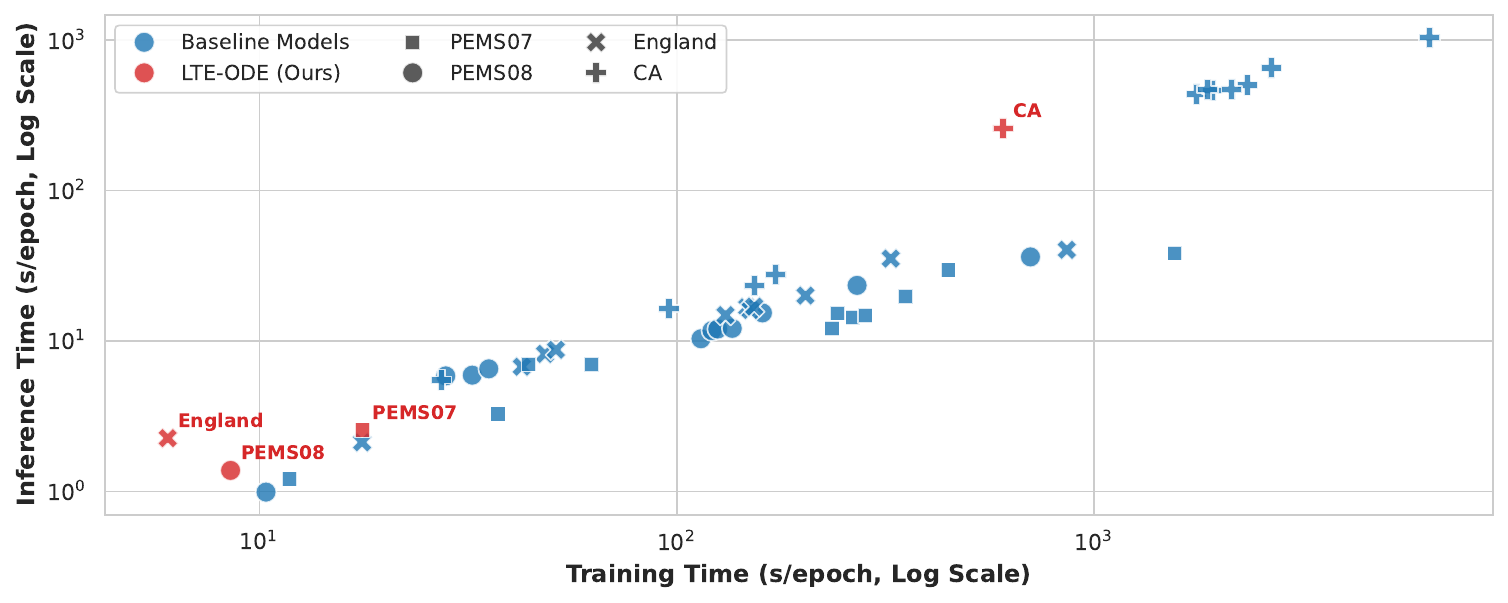}
		\vspace{-0.3cm}
		\caption{Performance Comparison of Training and Inference Time.}
		\label{fig:time}
	\end{figure}

	To rigorously evaluate the computational overhead, we mapped the training and inference times of various models on a logarithmic scale. As illustrated in the scatter plot (Figure~\ref{fig:time}), LTE-ODE achieves a remarkable balance between model expressiveness and computational footprint, distinctly occupying the Pareto optimal frontier (the bottom-left quadrant). Compared to heavily parameterized Transformer architectures (e.g., PDFormer, UniST) that suffer from quadratic spatial-temporal complexity, LTE-ODE yields nearly two orders of magnitude speedup (e.g., executing inference in only 1.38s/epoch on PEMS08). 

	Notably, while the absolute runtime naturally increases on larger-scale graphs like CA-due to the inherent sensitivity of the ODE solver's integration steps to massive graph operations-our model maintains robust scalability. Even on the massive CA dataset, LTE-ODE operates up to 4$\times$ and 10$\times$ faster than PDFormer and UniST, respectively. This demonstrates that our discrete compensator and LTE-guided masking mechanisms are computationally frugal, successfully integrating physical inductive biases without the prohibitive overhead typically associated with deep Neural ODEs.
	
	\subsection{Ablation Study}
	\label{subsec:ablation}
	
	To thoroughly investigate the effectiveness of the proposed Local Truncation Error (LTE)-guided mechanism and the system's deployment adaptability, we designed critical ablation variants. Table \ref{tab:ablation-components} and Table \ref{tab:ablation-steps} summarize the impact of architectural components and ODE integration steps on predictive performance across two representative datasets (see Appendix \ref{subsec:efficiency_and_ablation}.
	
	\begin{table}[!htbp]
		\centering
		\caption{Ablation study on architectural components and parameter efficiency.}
		\label{tab:ablation-components}	
		\vspace{-0.2cm}
		\resizebox{0.85\linewidth}{!}{
			\begin{tabular}{l|c|ccc|ccc}
				\toprule
				\multirow{2}{*}{Model Variant} & \multirow{2}{*}{Params (M)}
				& \multicolumn{3}{c|}{England} 
				& \multicolumn{3}{c}{PEMS08}  \\
				\cmidrule(lr){3-5} \cmidrule(lr){6-8} 
				& 
				& MAE & MAPE & RMSE 
				& MAE & MAPE & RMSE  \\			
				\midrule
				w/o LTE   &  0.54 &  2.99 & 4.28 & 6.34 & 14.69  & 9.70  & 22.37\\
				w/o Compensation & 0.27 & 3.05 & 4.49 & 6.49 & 15.31 & 10.07  & 22.92 \\
				w/o Mask  & 0.40    & 3.03  & 4.42  & 6.38 & 14.52 & 9.56 & 22.12 \\
				w/ Manifold Penalty & 0.40  & 3.02 & 4.37 & 6.35 & 14.65 & 9.86 & 22.25 \\
				\midrule
				LTE-ODE (Ours) & 0.40    & \textbf{2.95}  & \textbf{4.27}  & \textbf{6.25} & \textbf{14.56} & \textbf{9.41} & \textbf{22.15} \\
				\bottomrule
		\end{tabular}}
	\end{table}
	\vspace{-0.4cm}
	
	\paragraph{Impact of Architectural Components.}
	We evaluated variants of our model to isolate the contributions of the continuous-discrete synergy (Table \ref{tab:ablation-components}):
	
	\textit{w/o LTE (Removal of Truncation Error Calculation Logic):} Instead of utilizing the mathematically derived LTE as a forward inductive bias, this variant removes the physical error calculation logic, relying purely on standard learned neural layers to generate the mask. Notably, this inflates the parameter count to 0.54M (vs. our 0.40M) while degrading performance (e.g., RMSE increases from 6.25 to 6.34 on England). This verifies that the explicit physical discrepancy between solvers provides a superior, parameter-free supervisory signal compared to brute-force neural parameterization.
		
	\textit{w/o Compensation (Removal of Discrete Branch): }This variant strips away the independent discrete shock compensator $g_\phi(\cdot)$. As shown in Table \ref{tab:ablation-components}, while it achieves the lowest parameter footprint (0.27M), its performance drastically deteriorates across all metrics (MAE rises from 14.56 to 15.31 on PEMS08). This empirically validates our topological hypothesis: without the discrete branch, the continuous vector field is bound by Lipschitz continuity, structurally failing to fit non-linear, divergent traffic shocks.
		
\textit{w/o Mask (Degeneration to Standard ODE):} By removing the dynamic attention mask $\mathbf{M}_t$, the discrete compensation is uniformly applied across the spatial graph regardless of local shock conditions. Despite having the same parameter count as the full LTE-ODE, it suffers from severe over-smoothing, yielding sub-optimal predictions. 
		
\textit{w/ Manifold Penalty (Empirical Validation of Attention Collapse):} To rigorously validate Proposition \ref{prop:collapse} and our ``no-regularization'' paradigm, we evaluate this variant by forcing a trajectory smoothing constraint ($\mathcal{L}_{\text{mani}} = \lambda \|\mathbf{E}_t\|$). While this variant degrades predictive performance (e.g., RMSE rises to 22.25 on PEMS08), its macro-level average metrics partially mask a severe structural failure. 

	\begin{figure}[!htbp]
		\centering
		\includegraphics[width=0.85\linewidth]{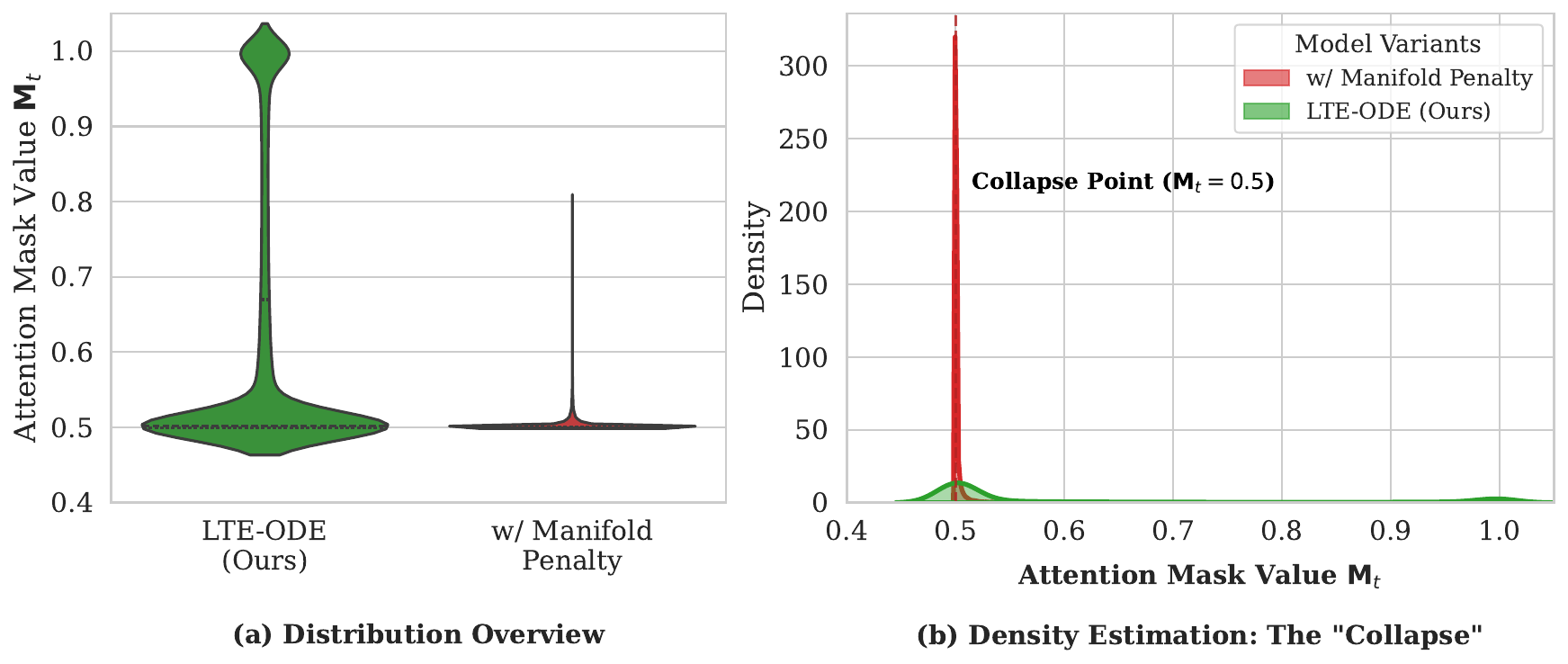}
		\vspace{-0.2cm}
		\caption{Spatial distribution of the attention mask $\mathbf{M}_t$, demonstrating Attention Collapse.}
		\label{fig:attention_collapse}
	\end{figure}
	
	The underlying pathology of the \textit{w/ Manifold Penalty} variant is unequivocally exposed in the internal inference dynamics (Figure \ref{fig:attention_collapse}). Penalizing the numerical discrepancy strips the model of its spatial selectivity, forcing the attention mask $\mathbf{M}_t$ to rigidly collapse into a Dirac-delta distribution at precisely $0.5$ (empirical mean $\approx 0.5007$). Consequently, the discrete compensator degenerates into uninformative spatial noise, blindly operating across all nodes regardless of local conditions. Conversely, our unpenalized LTE-ODE preserves a physically faithful heavy-tail distribution. It firmly anchors the $0.5$ baseline for smooth traffic while acutely activating towards $1.0$ the moment a non-linear shock is detected. Conversely, our unpenalized LTE-ODE preserves a physically faithful heavy-tail distribution-anchoring at $0.5$ for smooth traffic and activating towards $1.0$ during non-linear shocks (see Appendix \ref{app:smooth_prior} for theoretical details on this healthy physical prior).
	
	\begin{table}[!htbp]
		\caption{Sensitivity analysis of ODE integration steps against computational cost (FLOPs).}
		\vspace{-0.2cm}
		\label{tab:ablation-steps}
		\centering
		\resizebox{0.8\linewidth}{!}{
			\begin{tabular}{l|c|ccc|ccc}
				\toprule
				\multirow{2}{*}{Integration} 
				& \multirow{2}{*}{FLOPs} 
				& \multicolumn{3}{c|}{England} 
				& \multicolumn{3}{c}{PEMS08}  \\
				\cmidrule(lr){3-5} \cmidrule(lr){6-8} 
				Steps (1.0/$\Delta t$)  & (G) & MAE & MAPE & RMSE & MAE & MAPE & RMSE  \\
				\midrule
				\textit{steps=1} & \textbf{5.37}   & 3.06 & 4.53 & 6.40  & 14.87 & 9.72 & 22.52 \\
				\textit{steps=2} & 9.34  & 3.01 & 4.43  & 6.29 & \textbf{14.55} & 9.76  & \textbf{22.15}  \\
				\textit{\textbf{steps=4 (Ours)}} & 17.28  & \textbf{2.95}  & \textbf{4.27}  & \textbf{6.25} & 14.56 & \textbf{9.41} & \textbf{22.15} \\
				\textit{steps=6} & 17.45  & 3.05 & 4.54 & 6.47  & 15.10 & 9.76 & 22.71 \\
				\textit{steps=8}  & 33.17  & 3.01  & 4.41  & 6.38 & 14.64 & 9.51 & 22.21 \\
				\bottomrule
		\end{tabular}}
	\end{table}
	
	\paragraph{Integration Steps and Deployment Flexibility.}
	Table \ref{tab:ablation-steps} reveals a U-shaped performance curve regarding integration depth. Shallow integration (\textit{steps=1, 2}) slightly under-fits the complex continuous dynamics, while excessive steps (\textit{steps=6, 8}) accumulate numerical errors and double the computational overhead (FLOPs surge to 33.17G). Setting \textit{steps=4} achieves the optimal Pareto frontier for general performance. 
	Crucially, we elevate this hyperparameter from a mere tuning detail to a strategic lever for deployment flexibility. By adjusting the integration depth, LTE-ODE can seamlessly scale its computational footprint to match specific hardware constraints. For latency-sensitive edge computing environments (e.g., local traffic cameras or roadside sensor units with limited power), LTE-ODE can operate at \textit{steps=1} or \textit{steps=2}, dramatically reducing FLOPs while still maintaining competitive baseline performance. Conversely, for high-performance computing clusters (e.g., centralized city brains where precision is paramount), deploying with \textit{steps=4} unlocks the model's maximum predictive capacity. This demonstrates that LTE-ODE is not just theoretically sound, but highly adaptable for practical, large-scale industrial deployment.

In summary, the continuous ODE backbone, LTE-guided mask, and discrete compensator are indispensable to LTE-ODE. Together, they resolve the inherent tension between structural Lipschitz constraints and non-linear traffic volatility, achieving optimal robustness and efficiency (details in Appendix \ref{subsec:efficiency_and_ablation}).	
	
\section{Related Work}
\label{sec:related_work}
\vspace{-0.2cm}

\textbf{Deep Spatiotemporal Forecasting.}
Traffic forecasting is fundamentally challenging due to the complex, non-linear spatiotemporal dependencies inherent in road networks \citep{rahmani2023graph, tedjopurnomo2020survey}. Early approaches predominantly relied on statistical models like seasonal ARIMA \citep{williams2003modeling} and shallow machine learning techniques \citep{drucker1996support}. The paradigm subsequently shifted towards deep learning, with CNNs and RNNs being deployed to capture spatial proximity and temporal trends \citep{zhang2017deep, shi2015convolutional, liu2019acfm}.
Recognizing the non-Euclidean nature of transportation systems, Spatial-Temporal Graph Neural Networks \citep{song2020spatial}  have emerged as the mainstream solution. Building upon standard graph convolutions \citep{kipf2016semi}, foundational models \citep{yu2017spatio, guo2019attention} typically employed pre-defined graphs. To circumvent the reliance on priors, subsequent research introduced adaptive graph generation \citep{bai2020adaptive} and dynamic graph structure learning \citep{wu2020connecting, han2021dynamic} to capture hidden, time-varying correlations. 
As urban datasets expanded to citywide scales \citep{liu2023largest}, the quadratic computational complexity of spatial message passing became a critical bottleneck. Consequently, recent efforts have focused on enhancing scalability through efficient Transformer variants \citep{fang2025efficient,wang2023pfnet, GSNet,UniST, GPT-ST}. 
However, these advanced architectures predominately operate on discrete time snapshots. This inherent discreteness limits their capability to natively handle asynchronous spatial dependencies and irregular traffic time series \citep{zhang2024irregular}, failing to mirror the fundamentally continuous evolution of real-world traffic flows.

\textbf{Continuous-Time Dynamics and Neural ODEs.}
To address the limitations of discrete modeling, recent literature has begun conceptualizing traffic prediction through continuous-time paradigms. For instance, neural point processes \citep{jin2023spatio} and marked graph processes \citep{zhang2024deep} have been introduced to forecast sparsely distributed traffic congestion events along a continuous time axis.
More systematically, Neural Ordinary Differential Equations (Neural ODEs) have provided a principled framework to parameterize the derivative of hidden states. Integrating ODEs with GNNs \citep{liu2025graph} has proven highly effective in capturing spatiotemporal dynamics synchronously, as demonstrated by STGODE \citep{fang2021spatial}. Building on this, \citet{long2024unveiling} explicitly modeled the time delay of spatial information propagation—a crucial but often overlooked traffic property \citep{PDFormer}-using spatial-temporal delay differential equations. Furthermore, continuous-time diffusion models \citep{ruhling2023dyffusion} and equivariant neural fields \citep{knigge2024space} have shown promise in learning complex partial differential equations. Despite these theoretical strides, scaling these continuous-time solvers to highly diverse and overdispersed urban street networks \citep{kaiser2025spatio} remains computationally daunting and prone to unstable gradient propagation. 
To stabilize continuous neural networks and ensure they adhere to real-world physical dynamics, Physics-Informed Neural Networks (PINNs) have been widely adopted. By embedding fluid dynamics or partial differential equations as physical priors, models can achieve better interpretability and robustness in strict physical environments \citep{wilkman2025online,meng2023pinn,li2025embedding}. In standard physics-informed continuous learning, a common practice is to apply manifold regularization-specifically, penalizing the numerical discrepancy or local truncation error to force the evolutionary trajectory to remain strictly smooth and physically compliant.


In contrast, LTE-ODE abandons the assumption of strict manifold smoothness. Instead of minimizing the numerical solver discrepancy, we innovatively repurpose it as an unsupervised forward inductive bias. By using the LTE to selectively drive a spatial attention mask, our method bridges macroscopic continuous rhythms and microscopic discrete mutations, elegantly resolving the limitations of applying rigid physical constraints to stochastic traffic flows.
	
	\section{Conclusion}
	\label{sec:conclusion}
	\vspace{-0.2cm}
	In this paper, we propose LTE-ODE, a novel continuous-discrete hybrid dynamical system that fundamentally resolves the inherent conflict between smooth macroscopic rhythms and abrupt anomalies in spatiotemporal forecasting. Innovatively repurposing the Local Truncation Error (LTE) from a parallel dual-solver as an unsupervised structural signal, our framework dynamically triggers a discrete compensation branch to instantly break Lipschitz topological constraints at localized shock points. Extensive experiments on four large-scale real-world datasets confirm that LTE-ODE not only achieves state-of-the-art predictive accuracy in capturing highly non-linear traffic fluctuations, but also exhibits exceptional computational efficiency and hardware-aware deployment flexibility.
		
	{
		\small
		\bibliographystyle{plainnat}
		\bibliography{myref}
	}
	
	\newpage
	\appendix
	
	\section{Mathematical Proofs}
	\label{app:proofs}
	
	\subsection{Proof of Proposition \ref{prop:topology}}
	\label{app:proof_topology}

	\subsection{Proof of Proposition \ref{prop:collapse}}
	\label{app:proof_collapse}

	\subsection{Proof of Proposition \ref{prop:complexity}}
	\label{app:proof_complexity}

	\section{Experimental Details}
	\label{app:exp_details}

	\subsection{Efficiency, Scalability, and Component Analysis}
	\label{subsec:efficiency_and_ablation}

	\subsubsection{Discussion on the Smooth Prior vs. Artificial Collapse.}
	\label{app:smooth_prior}

		\section{Scope and Limitations}
	\label{sec:limitations}

\end{document}